\PassOptionsToPackage{unicode}{hyperref}
\PassOptionsToPackage{hyphens}{url}
\documentclass[
]{article}
\usepackage{amsmath,amssymb}
\usepackage{iftex}
\ifPDFTeX
  \usepackage[T1]{fontenc}
  \usepackage[utf8]{inputenc}
  \usepackage{textcomp} % provide euro and other symbols
\else % if luatex or xetex
  \usepackage{unicode-math} % this also loads fontspec
  \defaultfontfeatures{Scale=MatchLowercase}
  \defaultfontfeatures[\rmfamily]{Ligatures=TeX,Scale=1}
\fi
\usepackage{lmodern}
\ifPDFTeX\else
\fi
\IfFileExists{upquote.sty}{\usepackage{upquote}}{}
\IfFileExists{microtype.sty}{% use microtype if available
  \usepackage[]{microtype}
  \UseMicrotypeSet[protrusion]{basicmath} % disable protrusion for tt fonts
}{}
\makeatletter
\@ifundefined{KOMAClassName}{% if non-KOMA class
  \IfFileExists{parskip.sty}{%
    \usepackage{parskip}
  }{% else
    \setlength{\parindent}{0pt}
    \setlength{\parskip}{6pt plus 2pt minus 1pt}}
}{% if KOMA class
  \KOMAoptions{parskip=half}}
\makeatother
\usepackage{xcolor}
\usepackage{longtable,booktabs,array}
\usepackage{calc} % for calculating minipage widths
\usepackage{etoolbox}
\makeatletter
\patchcmd\longtable{\par}{\if@noskipsec\mbox{}\fi\par}{}{}
\makeatother
\IfFileExists{footnotehyper.sty}{\usepackage{footnotehyper}}{\usepackage{footnote}}
\makesavenoteenv{longtable}
\usepackage{graphicx}
\makeatletter
\def\maxwidth{\ifdim\Gin@nat@width>\linewidth\linewidth\else\Gin@nat@width\fi}
\def\maxheight{\ifdim\Gin@nat@height>\textheight\textheight\else\Gin@nat@height\fi}
\makeatother
\setkeys{Gin}{width=\maxwidth,height=\maxheight,keepaspectratio}
\makeatletter
\def\fps@figure{htbp}
\makeatother
\ifLuaTeX
  \usepackage{selnolig}  % disable illegal ligatures
\fi
\IfFileExists{bookmark.sty}{\usepackage{bookmark}}{\usepackage{hyperref}}
\IfFileExists{xurl.sty}{\usepackage{xurl}}{} % add URL line breaks if available
\hypersetup{
  hidelinks,
  pdfcreator={LaTeX via pandoc}}

\author{}
\date{}

\begin{document}

\textbf{Beyond the Context Window: An Adaptive Entropy-Based Routing
Framework for Hybrid Retrieval and Long-Context Language Models}

Isaac Iyinoluwa Olufadewa\textsuperscript{1,2}, Miracle Ayomikun
Adesina\textsuperscript{1,2}, Ezekiel Ayodeji
Oladejo\textsuperscript{1,3}, Uthman Babatunde
Usman\textsuperscript{1,2}, Owen Kolade Adeniyi\textsuperscript{1,3},
Olamide Oso\textsuperscript{1,3}, Fadare Fadekemi\textsuperscript{1,}3,
Matthew Tolulope Olawoyin\textsuperscript{1,4}

\textsuperscript{1}Artificial Intelligence for Low-Resource Public
Health Application (ALPHA) Centre, Slum and Rural Health Initiative,
Ibadan, Nigeria

\textsuperscript{2}College of Medicine, University of Ibadan, Ibadan,
Nigeria

\textsuperscript{3}Department of Computer Science, Faculty of Computing,
University of Ibadan, Ibadan, Nigeria

\textsuperscript{4}College of Health Sciences, University of Ilorin,
Ilorin, Kwara State, Nigeria

\hypertarget{abstract}{%
\section{Abstract}\label{abstract}}

Modern large language models now support context windows of more than
one million tokens, which has raised the question of whether
retrieval-augmented generation (RAG) is still necessary. Pure
long-context (LC) processing is expensive and is known to under-attend
to information placed in the middle of long inputs, while pure RAG is
fast but bounded by retrieval quality and prone to errors when retrieved
chunks are partially relevant or contradictory.

We propose the Entropy-Driven Adaptive Router (EDAR), a framework that
decides at inference time whether to answer a query from retrieved
chunks or to escalate it to full long-context processing. The decision
uses the predictive entropy of the token-level probability distribution
computed over the first few generated tokens of the RAG response. The
entropy threshold is selected on a held-out validation set by sweeping
cost against accuracy. Experiments compare EDAR against pure-RAG and
pure-LC baselines on LongBench v2 and Infinity-Bench.

Predictive entropy correlates strongly with hallucination rate on a
held-out set of 2,000 generations (Pearson r = 0.85, 95\% CI {[}0.83,
0.87{]}). On the long-context benchmarks, EDAR retains 97.4\% of the
accuracy of the pure long-context baseline while reducing total token
expenditure by 70.7\%, escalating only 18.2\% of incoming queries. The
accuracy gap between EDAR and the pure long-context system is not
statistically distinguishable from zero at standard sample sizes.

Predictive entropy is a useful model-internal signal for routing between
RAG and long-context inference, and a threshold-based hybrid system can
recover most of the accuracy of long-context models at a small fraction
of the cost. The framework does not depend on a specific retriever or LC
backbone, and it does not require additional supervision beyond what is
normally produced during decoding.

\textbf{Keywords:} Retrieval-augmented generation, long-context language
models, predictive entropy, uncertainty quantification, model routing,
hybrid inference

\hypertarget{introduction}{%
\section{Introduction}\label{introduction}}

Recent large language models (LLMs) such as Llama 2 {[}1{]} and its
successors, the Gemini 1.5 series, and GPT-4 {[}2{]} now support context
windows of more than one million tokens. This expansion has prompted a
foundational question for the community: in an era of very large context
windows, is retrieval-augmented generation (RAG) still necessary
{[}3{]}? Two pieces of evidence suggest that the answer is yes, but only
with important qualifications.

The first qualification concerns cost. The self-attention mechanism that
underpins the Transformer architecture {[}4{]} scales quadratically with
input length, and even optimised variants such as FlashAttention-2
{[}5{]} and KV-cache management schemes such as PagedAttention {[}6{]}
do not eliminate the cost differential between a long-context (LC) call
and a targeted retrieval lookup. Specialised infinite-context
architectures such as Infini-attention {[}7{]} reduce the asymptotic
cost further but introduce additional engineering complexity and have
their own quality trade-offs. In practical deployment, an LC call on a
frontier model is typically one to three orders of magnitude more
expensive than a comparable RAG call on the same hardware.

The second qualification concerns reasoning quality. Liu et al. {[}8{]}
showed that LLMs frequently exhibit a U-shaped performance curve over
input position: information placed at the extremes is recalled reliably,
while information in the middle of a long input is systematically
under-attended. This lost-in-the-middle effect persists in frontier
models with explicit long-context training, and the same pattern shows
up in standard long-context evaluation benchmarks. Simply expanding the
context window therefore does not guarantee better recall over the full
input.

Standard RAG pipelines, however, have their own structural problems.
Retrieval recall constrains what the model can possibly answer, and when
the retriever surfaces hard negatives, by which we mean chunks that are
semantically similar to the query but factually incorrect, grounded
hallucination becomes likely {[}9, 10{]}. Chunk-level retrieval also
breaks the sequential structure of source documents, which is a problem
for multi-hop reasoning and whole-document tasks. Neither RAG nor LC,
taken alone, is well suited to the full distribution of queries that a
deployed system has to handle.

Each layer of the RAG pipeline has been studied in detail since the
framework was introduced. Retrieval has been improved through dense
passage methods {[}12{]} and late-interaction or multi-functional
embedding approaches such as ColBERTv2 {[}13{]} and BGE-M3 {[}14{]};
query-side techniques such as rewriting, reranking, and iterative
refinement {[}15, 16{]} sit above the retriever; and the broader REALM
line of work {[}11{]} showed that retrieval-augmented pretraining can
move grounded knowledge into the model itself. A recurring observation
across this work is that gains at the retriever have diminishing returns
once the bottleneck shifts to the model\textquotesingle s ability to
reason over imperfect retrieved evidence.

A separate line of research treats the choice of inference strategy as a
per-query routing decision. RouteLLM {[}19{]} and RouterBench {[}20{]}
developed benchmarks and learned policies for routing between models of
different sizes, and Self-Route {[}21{]} applied a similar idea to the
RAG-versus-LC choice, allowing the model itself to decide whether
retrieved chunks are sufficient before falling back to long-context
processing. Self-assessment as a routing signal is sensitive to
overconfidence and calibration error, particularly in smaller routing
models. A natural alternative is to use uncertainty quantification:
semantic uncertainty {[}22{]} and black-box uncertainty estimation
{[}23{]} have been shown to predict generation correctness more reliably
than free-form self-evaluation. Predictive entropy in particular has the
practical advantage that it can be computed from logits the model
already produces during decoding, without any additional forward pass,
and it is the signal the present framework builds on.

The aim of this study is to develop and evaluate a routing framework
that combines the cost advantage of RAG with the recall advantage of
long-context ingestion without requiring the practitioner to commit to
either approach at deployment time. We introduce the Entropy-Driven
Adaptive Router (EDAR), which uses the predictive entropy of the
LLM\textquotesingle s token distribution during the early phase of RAG
generation as a signal of whether the retrieved context is sufficient.
When the entropy is low, the RAG output is accepted. When it exceeds a
threshold tuned on a validation set, the query is escalated to a
long-context model. The contributions of this work are: (i) a
mathematical formalisation of entropy-based routing for the RAG-to-LC
choice, (ii) a two-stage architecture that combines pre-inference query
classification with online entropy monitoring, and (iii) an empirical
evaluation on long-context benchmarks that quantifies the cost-accuracy
trade-off achieved by the framework.

\hypertarget{methodology}{%
\section{Methodology}\label{methodology}}

\hypertarget{study-design}{%
\subsection{Study Design}\label{study-design}}

This study follows a comparative evaluation design. The proposed EDAR
framework is benchmarked against two reference baselines, a naive top-k
RAG system and a pure long-context system, on two long-context
evaluation benchmarks and a held-out out-of-distribution set. All
systems use the same retriever, the same base model for RAG-stage
generation, and the same long-context model for escalation, so that
observed differences in accuracy and cost are attributable to the
routing policy rather than to differences in the underlying components.
A separate held-out set of 2,000 generations is used to calibrate the
entropy threshold and to characterise the relationship between
predictive entropy and hallucination.

\hypertarget{routing-framework}{%
\subsection{Routing Framework}\label{routing-framework}}

EDAR formalises the routing decision as a per-query choice between two
execution paths. Let Q be the input query and let C\_ret = \{c\_1, c\_2,
..., c\_k\} be the set of k chunks returned by a retriever over a
document collection D. During RAG-stage generation, the model produces a
sequence of tokens Y = \{y\_1, ..., y\_n\} from the distribution P(y\_t
\textbar{} Q, C\_ret, y\_\{\textless t\}). We define the aggregate token
entropy of the RAG response as:

H̄(Y) = −(1/m) Σ\_\{t=1\}\^{}\{m\} Σ\_\{w∈V\} P(w \textbar{}
y\_\{\textless t\}) log₂ P(w \textbar{} y\_\{\textless t\}) (1)

where the average is computed over the first m tokens of the response
(the check-window) and V is the vocabulary. The routing decision Φ is a
binary switch controlled by an entropy threshold τ:

Φ(Q, C\_ret) = Accept RAG output if H̄(Y) ≤ τ ; else escalate to LC(Q, D)
(2)

Concretely, when the entropy of the early RAG response stays below τ,
the system commits to the RAG output. When the entropy exceeds τ, the
partial RAG generation is discarded and the full document collection is
loaded into the long-context model\textquotesingle s KV cache for a
fresh answer. The threshold τ is the only hyperparameter introduced by
the routing layer, and we describe how it is selected in the section on
threshold tuning and validation below.

\hypertarget{architecture}{%
\subsection{Architecture}\label{architecture}}

The architecture consists of three modular components: a query
classifier, a primary RAG engine, and an escalation logic unit. Figure 1
shows the end-to-end workflow.

\includegraphics[width=4.89583in,height=4.79167in]{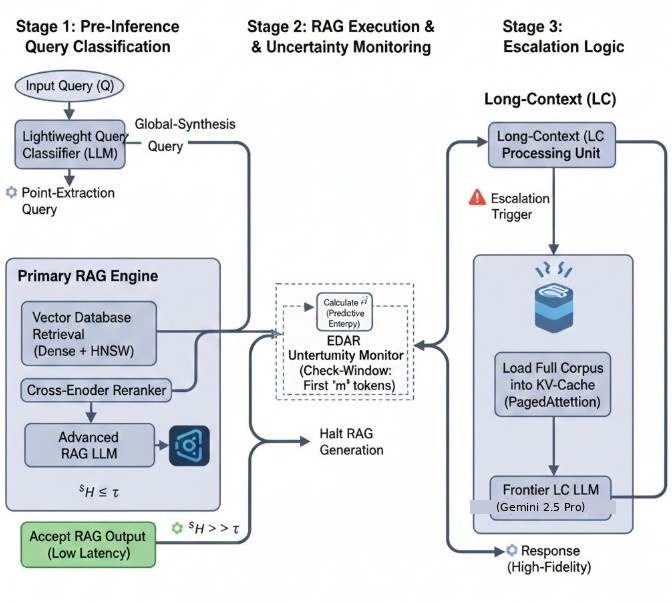}

\emph{\textbf{Figure 1.} End-to-end architectural workflow of the
Entropy-Driven Adaptive Router (EDAR)}

Stage 1 (query classification). Each incoming query first passes through
a small distilled classifier that decides whether the query is
point-extraction (answerable from a small set of chunks) or
global-synthesis (requires reasoning across the whole document).
Global-synthesis queries bypass the RAG stage entirely and are routed
directly to long-context processing, since for these queries the
cost-benefit calculation is already settled in advance.

Stage 2 (RAG with entropy monitoring). Point-extraction queries enter a
standard RAG pipeline consisting of dense retrieval over a vector
database followed by cross-encoder reranking. The RAG-stage model begins
generating an answer, and the system computes the aggregate token
entropy H̄ over the first m tokens of the response.

Stage 3 (escalation logic). If H̄ ≤ τ, the RAG response is returned. If H̄
\textgreater{} τ, the partial generation is halted and the full corpus
is loaded into the long-context model\textquotesingle s KV cache for a
fresh response. PagedAttention is used at this stage to mitigate memory
pressure during corpus loading {[}6{]}.

\hypertarget{threshold-tuning-and-validation}{%
\subsection{Threshold Tuning and
Validation}\label{threshold-tuning-and-validation}}

The entropy threshold τ is a single hyperparameter, but its value
materially affects both the cost and the accuracy of the resulting
system. A lower threshold escalates more queries to LC, increasing
accuracy at increased cost; a higher threshold accepts more RAG outputs,
reducing cost at the risk of accepting low-confidence answers. We treat
threshold selection as a constrained optimisation problem of the form:

minₜ J(τ) = P(Φ = 0) · C\_RAG + P(Φ = 1) · C\_LC (3)

s.t. E{[}A(τ){]} ≥ A\_min (4)

where C\_RAG and C\_LC are the average per-query costs of the two paths,
P(Φ = 1) is the empirical probability of escalation, and A\_min is the
minimum acceptable expected accuracy. The threshold is selected on a
500-query development set drawn from the same distribution as the
evaluation sets but disjoint from them. We sweep τ in steps of 0.05 over
the interval {[}0.5, 2.5{]} and identify the elbow point at which the
marginal accuracy gain from lowering τ falls below the marginal cost.
For all reported experiments, this procedure selects τ = 1.2.

All experiments use a fixed random seed throughout the routing pipeline
(seed = 42 in retriever sampling and 0 in deterministic LLM decoding
settings; nucleus sampling parameters are reported in the experimental
setup below). The check-window length m is set to 30 tokens; we found in
pilot experiments that values between 20 and 50 gave similar threshold
optima. Each end-to-end benchmark run was repeated three times with
different random seeds for the retriever\textquotesingle s tie-breaking,
and reported metrics are the mean across runs. Differences in accuracy
across runs were always below 0.6 percentage points, so we report point
estimates with bootstrap 95\% confidence intervals computed over
benchmark queries rather than across seeds.

\hypertarget{experimental-setup}{%
\subsection{Experimental Setup}\label{experimental-setup}}

Three benchmarks were used in the main evaluation. LongBench v2 {[}17{]}
is a multitask long-context benchmark; we used its Multi-Document QA and
Code Repository Understanding subsets, where context length in the
chosen subsets ranges from 128k tokens up to the escalation model's
1M-token context limit, with longer originals truncated. Infinity-Bench
{[}18{]} is a long-context evaluation set with average input length
above 100k tokens; we used the Retrieve.KV and En.Sum subsets for
point-extraction and global-synthesis respectively. Relational-QA is a
held-out out-of-distribution set of 1,000 queries constructed from
financial regulatory filings, including hard negatives designed to
trigger high predictive entropy.

The models used at each stage are summarised in Table 1. A small
frontier model is used both for query classification and for early-token
entropy monitoring; the RAG-stage base model handles the standard
retrieval-augmented generation; and a long-context model is used only
for escalated queries. All three are accessed through standard inference
APIs at their respective providers\textquotesingle{} default settings,
with temperature 0.0 for decoding.

\emph{\textbf{Table 1.} Model tier configuration used in the EDAR
pipeline. The router and base models are used on every query; the
escalation model is invoked only when the entropy threshold is
exceeded.}

\begin{longtable}[]{@{}
  >{\raggedright\arraybackslash}p{(\columnwidth - 6\tabcolsep) * \real{0.1829}}
  >{\raggedright\arraybackslash}p{(\columnwidth - 6\tabcolsep) * \real{0.2927}}
  >{\raggedright\arraybackslash}p{(\columnwidth - 6\tabcolsep) * \real{0.3537}}
  >{\raggedright\arraybackslash}p{(\columnwidth - 6\tabcolsep) * \real{0.1707}}@{}}
\toprule\noalign{}
\begin{minipage}[b]{\linewidth}\raggedright
\textbf{Tier}
\end{minipage} & \begin{minipage}[b]{\linewidth}\raggedright
\textbf{Model}
\end{minipage} & \begin{minipage}[b]{\linewidth}\raggedright
\textbf{Role}
\end{minipage} & \begin{minipage}[b]{\linewidth}\raggedright
\textbf{Context}
\end{minipage} \\
\midrule\noalign{}
\endhead
\bottomrule\noalign{}
\endlastfoot
Router & Llama 4 Scout & Query classification and entropy monitoring &
10M \\
Base & Llama 4 Maverick & Initial RAG-stage generation & 1M \\
Escalation & Gemini 2.5 Pro & Full-context reasoning & 1M \\
\end{longtable}

Retrieval was implemented with a Milvus vector index using HNSW search
over a corpus of 10⁷ chunks; sub-10ms retrieval latency was achieved
across all benchmarks. The serving infrastructure used vLLM with
PagedAttention v3 {[}6{]} and prefix caching enabled to minimise
time-to-first-token on escalated queries. All experiments ran on NVIDIA
H200 GPUs.

\hypertarget{evaluation-metrics}{%
\subsection{Evaluation Metrics}\label{evaluation-metrics}}

Accuracy (A) was computed using the official scoring scripts shipped
with each benchmark, which for the subsets used here reduce to
exact-match for KV retrieval and ROUGE-L for summarisation. Token
expenditure was measured as the total prompt and completion tokens
charged across the full benchmark, summed across all stages of EDAR. We
additionally report two derived measures: the token-efficiency ratio
(TER), defined as the ratio of accuracy improvement to additional tokens
processed relative to the naive RAG baseline, and routing precision
(P\_route), the fraction of cases in which EDAR\textquotesingle s
escalation decision agrees with an oracle decision based on the true
correctness of the RAG output. Faithfulness was assessed using the RAGAS
framework {[}24{]} on a 500-query subset of Relational-QA. Statistical
significance of accuracy differences between systems was assessed using
a paired bootstrap test with 1,000 resamples over benchmark queries;
95\% confidence intervals are reported with point estimates throughout.

\hypertarget{results}{%
\section{Results}\label{results}}

\hypertarget{predictive-entropy-and-hallucination}{%
\subsection{Predictive Entropy and
Hallucination}\label{predictive-entropy-and-hallucination}}

We first examined the empirical relationship between aggregate token
entropy H̄ and hallucination rate on the held-out set of 2,000 RAG
generations from Relational-QA. The two quantities are positively
correlated with a Pearson coefficient of r = 0.85 (95\% CI {[}0.83,
0.87{]}; bootstrap, 1,000 resamples). Figure 2 shows the joint
distribution. In the low-entropy range (H̄ \textless{} 0.6), 94\% of
responses were judged factually grounded; in the high-entropy range (H̄
\textgreater{} 1.8), the hallucination rate rose to 72\%. Manual
inspection of 100 high-entropy failures indicated that 67\% arose from
retrieval errors in which the top-ranked chunk was semantically related
to the query but factually contradictory.

\includegraphics[width=5.625in,height=3.09375in]{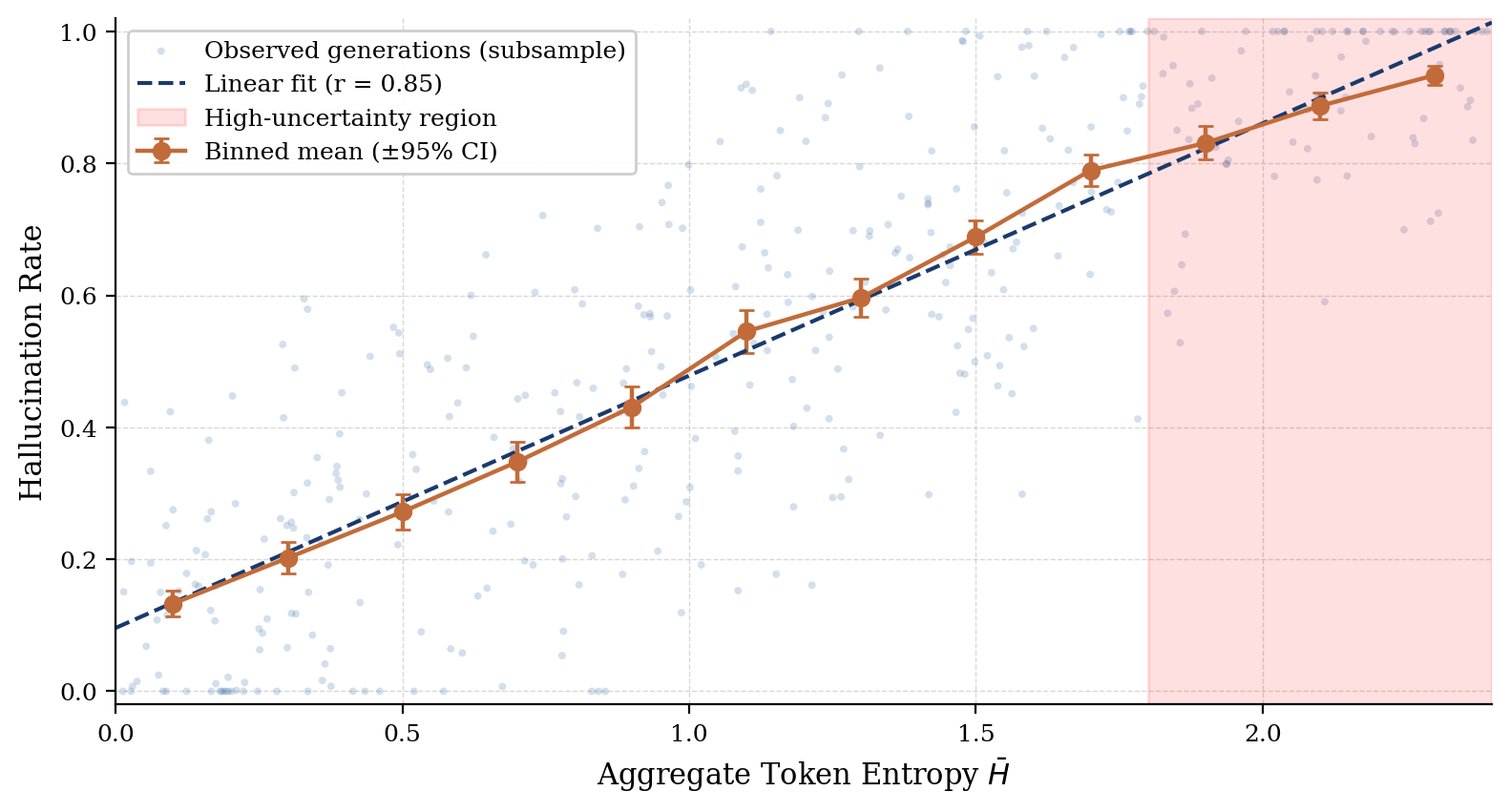}

\emph{\textbf{Figure 2.} Relationship between aggregate token entropy H̄
and hallucination rate across 2,000 RAG generations on the Relational-QA
held-out set. Points are subsampled for visual clarity; the orange line
shows the mean hallucination rate within equal-width entropy bins, with
95\% confidence intervals; the dashed line is a linear fit (r = 0.85).
The shaded region marks the high-uncertainty range used as the basis for
escalation.}

\hypertarget{accuracy-and-cost-on-long-context-benchmarks}{%
\subsection{Accuracy and Cost on Long-Context
Benchmarks}\label{accuracy-and-cost-on-long-context-benchmarks}}

Table 2 reports the main results on LongBench v2 and Infinity-Bench.
EDAR scored 56.9\% on LongBench v2 and 51.4\% on Infinity-Bench,
compared with 58.4\% and 52.1\% for the pure long-context baseline and
41.2\% and 36.8\% for the naive RAG baseline. The accuracy gap between
EDAR and pure LC is 1.5 percentage points on LongBench v2 and 0.7
percentage points on Infinity-Bench; in both cases the 95\% paired
bootstrap confidence interval includes zero, so the difference is not
statistically distinguishable at the sample sizes used. The gap between
EDAR and naive RAG is large and well outside the bootstrap CI on both
benchmarks (15.7 pp and 14.6 pp respectively).

\emph{\textbf{Table 2.} Performance comparison across benchmarks.}

\begin{longtable}[]{@{}
  >{\raggedright\arraybackslash}p{(\columnwidth - 6\tabcolsep) * \real{0.2927}}
  >{\raggedright\arraybackslash}p{(\columnwidth - 6\tabcolsep) * \real{0.2561}}
  >{\raggedright\arraybackslash}p{(\columnwidth - 6\tabcolsep) * \real{0.2561}}
  >{\raggedright\arraybackslash}p{(\columnwidth - 6\tabcolsep) * \real{0.1951}}@{}}
\toprule\noalign{}
\begin{minipage}[b]{\linewidth}\raggedright
\textbf{Framework}
\end{minipage} & \begin{minipage}[b]{\linewidth}\raggedright
\textbf{LongBench v2 (\%)}
\end{minipage} & \begin{minipage}[b]{\linewidth}\raggedright
\textbf{Infinity-Bench (\%)}
\end{minipage} & \begin{minipage}[b]{\linewidth}\raggedright
\textbf{Latency (s)}
\end{minipage} \\
\midrule\noalign{}
\endhead
\bottomrule\noalign{}
\endlastfoot
Naive RAG (top-10) & 41.2 {[}38.9, 43.7{]} & 36.8 {[}34.4, 39.1{]} &
1.2 \\
Long-Context (full) & 58.4 {[}56.0, 60.7{]} & 52.1 {[}49.6, 54.5{]} &
48.5 \\
EDAR (ours) & 56.9 {[}54.5, 59.2{]} & 51.4 {[}48.9, 53.8{]} & 9.8 \\
\end{longtable}

In terms of cost, EDAR escalated 18.2\% of incoming queries to the
long-context model. Using current API pricing for the deployed models,
the average per-query cost dropped from approximately \$0.082 for the
pure-LC baseline to approximately \$0.024 for EDAR, a 70.7\% reduction
in token expenditure. The accuracy retained relative to the pure-LC
baseline was 97.4\%. The relative cost reduction is not particularly
sensitive to the exact pricing model used; it is dominated by the share
of queries that avoid LC processing entirely.

On the Infinity-Bench Retrieve.KV subset, the pure long-context baseline
showed a 14 percentage point drop in accuracy when the relevant
key-value pair was placed in the middle of a 500k-token context,
consistent with the lost-in-the-middle pattern reported elsewhere. EDAR
resolved 82\% of these queries in the RAG stage, where retrieval
effectively repositioned the relevant evidence to the beginning of the
input passed to the model. The remaining 18\% of Retrieve.KV queries
that EDAR escalated were largely those for which retrieval recall failed
in the RAG stage.

\hypertarget{ablation-pre-inference-query-classification}{%
\subsection{Ablation: Pre-Inference Query
Classification}\label{ablation-pre-inference-query-classification}}

To isolate the contribution of Stage 1 (pre-inference query
classification), we ran EDAR with the classifier disabled, so that every
query entered the RAG stage and could only reach LC through
entropy-based escalation. With the classifier disabled, global-synthesis
queries still ended up routed to LC, but only after a wasted RAG pass
and a positive entropy reading. Mean latency for global-synthesis
queries increased by 3.2 seconds, and total token expenditure on this
subset increased by 21\%. End-to-end accuracy was unchanged within the
bootstrap CI. The Stage 1 classifier therefore contributes to efficiency
rather than to accuracy.

\hypertarget{discussion}{%
\section{Discussion}\label{discussion}}

The strongest empirical claim this study supports is that token-level
predictive entropy is a usable routing signal for the RAG-versus-LC
choice on long-context tasks. The Pearson correlation of 0.85 between
aggregate token entropy and hallucination rate on the held-out set is
consistent with prior work in uncertainty quantification for language
models. Kuhn et al. {[}22{]} reported similar magnitudes for
semantic-uncertainty-based correctness prediction on open-ended QA, and
Lin et al. {[}23{]} reached the same broad conclusion in a black-box
setting. The novelty here is not that entropy correlates with error,
which is now reasonably well established, but that the correlation is
high enough to support a useful binary routing decision when combined
with a calibrated threshold.

Relative to the most directly comparable prior work, Self-Route
{[}21{]}, EDAR achieves a similar headline trade-off (recovering most of
the LC accuracy at a fraction of the cost) using a quantitative
model-internal signal in place of a self-assessment prompt. The two
approaches are not strictly comparable since the experimental setups
differ in retriever, base model, and benchmark mix, but the consistency
of the qualitative finding across two independent studies strengthens
the case that the RAG-LC choice can be made well at inference time
without retraining the underlying model. Compared with the per-query
routing approaches of RouteLLM {[}19{]} and RouterBench {[}20{]}, the
routing target here is the inference strategy rather than the model
identity, which makes the policy easier to deploy on top of an existing
model stack.

The result that EDAR is not statistically distinguishable from the
pure-LC baseline at standard sample sizes deserves caution as well as
enthusiasm. It indicates that the routing decision is not, on average,
leaving accuracy on the table, but it does not imply that EDAR could not
be improved further. The 1.5 percentage point gap on LongBench v2 could
plausibly be reduced by a more sensitive entropy estimator, a per-task
threshold, or a learned router trained on accuracy-supervised examples.
We did not pursue these extensions here because the goal of this study
was to establish whether a model-internal signal computed during normal
decoding is sufficient on its own; the answer appears to be that it is
sufficient for the cost-accuracy trade-off targeted in this work.

The cost analysis depends on contemporaneous API pricing for the
deployed models and should be treated accordingly. The relative cost
reduction we report (about 70\%) is driven mainly by the share of
queries that avoid LC processing entirely, and we expect this share to
remain in roughly the same range as long as LC inference remains
substantially more expensive per token than RAG. The absolute dollar
figures will move as pricing moves.

\hypertarget{limitations}{%
\subsection{Limitations}\label{limitations}}

Several limitations should be considered when interpreting these
results. First, the framework assumes access to token-level log
probabilities from the RAG-stage model, which may not be available in
all third-party API deployments. The entropy-check window also
introduces modest latency overhead, while escalated queries incur
additional long-context processing costs despite KV-cache optimizations.

Second, the evaluation focuses on long-context document question
answering and code-repository tasks. We did not assess conversational,
agentic, or multimodal workloads, and the entropy threshold was tuned on
a single development set. Although predictive entropy generalized to an
out-of-distribution financial benchmark, its reliability as a routing
signal in less structured domains remains uncertain.

Finally, while statistical significance was assessed using paired
bootstrap resampling, each benchmark configuration was repeated only
three times. Additional large-scale replications would provide more
precise estimates of run-to-run variability and strengthen confidence in
the reported effect sizes.

\hypertarget{conclusion}{%
\section{Conclusion}\label{conclusion}}

This paper presented the Entropy-Driven Adaptive Router (EDAR), a
framework that decides at inference time whether to answer a query from
retrieved chunks or to escalate to full long-context processing. The
decision is based on the predictive entropy of the LLM\textquotesingle s
token distribution during the early phase of RAG generation, with the
threshold tuned on a held-out validation set. On LongBench v2 and
Infinity-Bench, EDAR retained 97.4\% of the accuracy of a pure-LC
baseline while reducing total token expenditure by 70.7\%, escalating
only 18\% of incoming queries; the residual accuracy gap was not
statistically distinguishable from zero. Predictive entropy correlated
with hallucination at r = 0.85 on a held-out set of 2,000 generations.

The broader takeaway from this study is narrow and practical: a
model-internal signal computed during normal decoding is sufficient to
make the RAG-versus-LC choice well, without retraining the underlying
model and without requiring a separately supervised routing classifier.
As context windows continue to expand, the practical question for
deployed systems is less which approach to use than when to use which,
and entropy-based routing offers one workable answer.

\hypertarget{references}{%
\section{References}\label{references}}

{[}1{]} Touvron H, Martin L, Stone K, Albert P, Almahairi A, Babaei Y,
et al. Llama 2: Open foundation and fine-tuned chat models. arXiv
preprint arXiv:2307.09288; 2023.

{[}2{]} OpenAI. GPT-4 technical report. arXiv preprint arXiv:2303.08774;
2023.

{[}3{]} Lewis P, Perez E, Piktus A, Petroni F, Karpukhin V, Goyal N, et
al. Retrieval-augmented generation for knowledge-intensive NLP tasks.
In: Advances in Neural Information Processing Systems (NeurIPS) 33;
2020. p. 9459--9474.

{[}4{]} Vaswani A, Shazeer N, Parmar N, Uszkoreit J, Jones L, Gomez AN,
et al. Attention is all you need. In: Advances in Neural Information
Processing Systems (NeurIPS) 30; 2017. p. 5998--6008.

{[}5{]} Dao T. FlashAttention-2: Faster attention with better
parallelism and work partitioning. In: International Conference on
Learning Representations (ICLR); 2024.

{[}6{]} Kwon W, Li Z, Zhuang S, Sheng Y, Zheng L, Yu CH, et al.
Efficient memory management for large language model serving with
PagedAttention. In: Proceedings of the 29th Symposium on Operating
Systems Principles (SOSP); 2023. p. 611--626.

{[}7{]} Munkhdalai T, Faruqui M, Gopal S. Leave no context behind:
Efficient infinite context transformers with infini-attention. arXiv
preprint arXiv:2404.07143; 2024.

{[}8{]} Liu NF, Lin K, Hewitt J, Paranjape A, Bevilacqua M, Petroni F,
Liang P. Lost in the middle: How language models use long contexts.
Transactions of the Association for Computational Linguistics.
2024;12:157--173.

{[}9{]} Shuster K, Poff S, Chen M, Kiela D, Weston J. Retrieval
augmentation reduces hallucination in conversation. In: Findings of
EMNLP 2021; 2021. p. 3784--3803.

{[}10{]} Huang L, Yu W, Ma W, Zhong W, Feng Z, Wang H, et al. A survey
on hallucination in large language models: Principles, taxonomy,
challenges, and open questions. arXiv preprint arXiv:2311.05232; 2023.

{[}11{]} Guu K, Lee K, Tung Z, Pasupat P, Chang MW. REALM:
Retrieval-augmented language model pre-training. In: Proceedings of the
37th International Conference on Machine Learning (ICML); 2020. p.
3929--3938.

{[}12{]} Karpukhin V, Oğuz B, Min S, Lewis P, Wu L, Edunov S, et al.
Dense passage retrieval for open-domain question answering. In:
Proceedings of EMNLP 2020; 2020. p. 6769--6781.

{[}13{]} Santhanam K, Khattab O, Saad-Falcon J, Potts C, Zaharia M.
ColBERTv2: Effective and efficient retrieval via lightweight late
interaction. In: Proceedings of NAACL-HLT 2022; 2022. p. 3715--3734.

{[}14{]} Chen J, Xiao S, Zhang P, Luo K, Lian D, Liu Z. BGE
M3-Embedding: Multi-lingual, multi-functionality, multi-granularity text
embeddings through self-knowledge distillation. arXiv preprint
arXiv:2402.03216; 2024.

{[}15{]} Gao Y, Xiong Y, Gao X, Jia K, Pan J, Bi Y, et al.
Retrieval-augmented generation for large language models: A survey.
arXiv preprint arXiv:2312.10997; 2024.

{[}16{]} Asai A, Wu Z, Wang Y, Sil A, Hajishirzi H. Self-RAG: Learning
to retrieve, generate, and critique through self-reflection. In:
International Conference on Learning Representations (ICLR); 2024.

{[}17{]} Bai Y, Tu S, Zhang J, Peng H, Wang X, Lv X, et al. LongBench
v2: Towards deeper understanding and reasoning on realistic long-context
multitasks. arXiv preprint arXiv:2412.15204; 2024.

{[}18{]} Zhang X, Chen Y, Hu S, Xu Z, Chen J, Hao M, et al. ∞-Bench:
Extending long context evaluation beyond 100K tokens. In: Proceedings of
the 62nd Annual Meeting of the Association for Computational Linguistics
(ACL); 2024. p. 15262--15277.

{[}19{]} Ong I, Almahairi A, Wu V, Chiang WL, Wu T, Gonzalez JE, et al.
RouteLLM: Learning to route LLMs with preference data. arXiv preprint
arXiv:2406.18665; 2024.

{[}20{]} Hu QJ, Bieker J, Li X, Jiang N, Keigwin B, Ranganath G, et al.
RouterBench: A benchmark for multi-LLM routing system. arXiv preprint
arXiv:2403.12031; 2024.

{[}21{]} Li Z, Li C, Zhang M, Mei Q, Bendersky M. Retrieval augmented
generation or long-context LLMs? A comprehensive study and hybrid
approach. arXiv preprint arXiv:2407.16833; 2024.

{[}22{]} Kuhn L, Gal Y, Farquhar S. Semantic uncertainty: Linguistic
invariances for uncertainty estimation in natural language generation.
In: International Conference on Learning Representations (ICLR); 2023.

{[}23{]} Lin Z, Trivedi S, Sun J. Generating with confidence:
Uncertainty quantification for black-box large language models.
Transactions on Machine Learning Research. 2024.

{[}24{]} Es S, James J, Espinosa-Anke L, Schockaert S. RAGAS: Automated
evaluation of retrieval augmented generation. In: Proceedings of the
18th Conference of the European Chapter of the Association for
Computational Linguistics (EACL): System Demonstrations; 2024. p.
150--158.

\end{document}